\documentclass[10pt,twocolumn,letterpaper]{article}

\usepackage[pagenumbers]{wacv}  % To force page numbers, e.g. for an arXiv version

\usepackage{graphicx}
\usepackage{amsmath}
\usepackage{amssymb}
\usepackage{booktabs}

\usepackage{kotex}
\usepackage{listings}
\usepackage{adjustbox}
\usepackage[linesnumbered, ruled,vlined]{algorithm2e}  
\usepackage{amsmath}
\usepackage{multicol}
\usepackage{multirow}
\usepackage{svg}
\usepackage[pagebackref,breaklinks,colorlinks]{hyperref}
\usepackage{comment}

\usepackage[capitalize]{cleveref}
\crefname{section}{Sec.}{Secs.}
\Crefname{section}{Section}{Sections}
\Crefname{table}{Table}{Tables}
\crefname{table}{Tab.}{Tabs.}

\begin{document}

\title{PTQ4VM: Post-Training Quantization for Visual Mamba }

\author{Younghyun Cho$^{*}$ \quad Changhun Lee$^{*}$ \quad Seonggon Kim \quad Eunhyeok Park\\
Pohang University of Science and Technology (POSTECH), Republic of Korea \\
\texttt{\{yhcho97, changhun.lee,  sungonuni, eh.park\}@postech.ac.kr} \\
}

\maketitle
\renewcommand{\thefootnote}{\fnsymbol{footnote}}
\footnotetext[1]{ These authors contributed equally.}
\renewcommand{\thefootnote}{\arabic{footnote}}
%%%%%%%%% ABSTRACT
\begin{abstract}
Visual Mamba is an approach that extends the selective space state model, Mamba, to vision tasks. It processes image tokens sequentially in a fixed order, accumulating information to generate outputs. Despite its growing popularity for delivering high-quality outputs at a low computational cost across various tasks, Visual Mamba is highly susceptible to quantization, which makes further performance improvements challenging. Our analysis reveals that the fixed token access order in Visual Mamba introduces unique quantization challenges, which we categorize into three main issues: 1) token-wise variance, 2) channel-wise outliers, and 3) a long tail of activations. To address these challenges, we propose Post-Training Quantization for Visual Mamba (PTQ4VM), which introduces two key strategies: Per-Token Static (PTS) quantization and Joint Learning of Smoothing Scale and Step Size (JLSS). To the our best knowledge, this is the first quantization study on Visual Mamba. PTQ4VM can be applied to various Visual Mamba backbones, converting the pretrained model to a quantized format in under 15 minutes without notable quality degradation. Extensive experiments on large-scale classification and regression tasks demonstrate its effectiveness, achieving up to 1.83$\times$ speedup on GPUs with negligible accuracy loss compared to FP16. Our code is available at \url{https://github.com/YoungHyun197/ptq4vm}.
\end{abstract}

%%%%%%%%% BODY TEXT

\begin{figure}
\centering
\includegraphics[width=\linewidth]{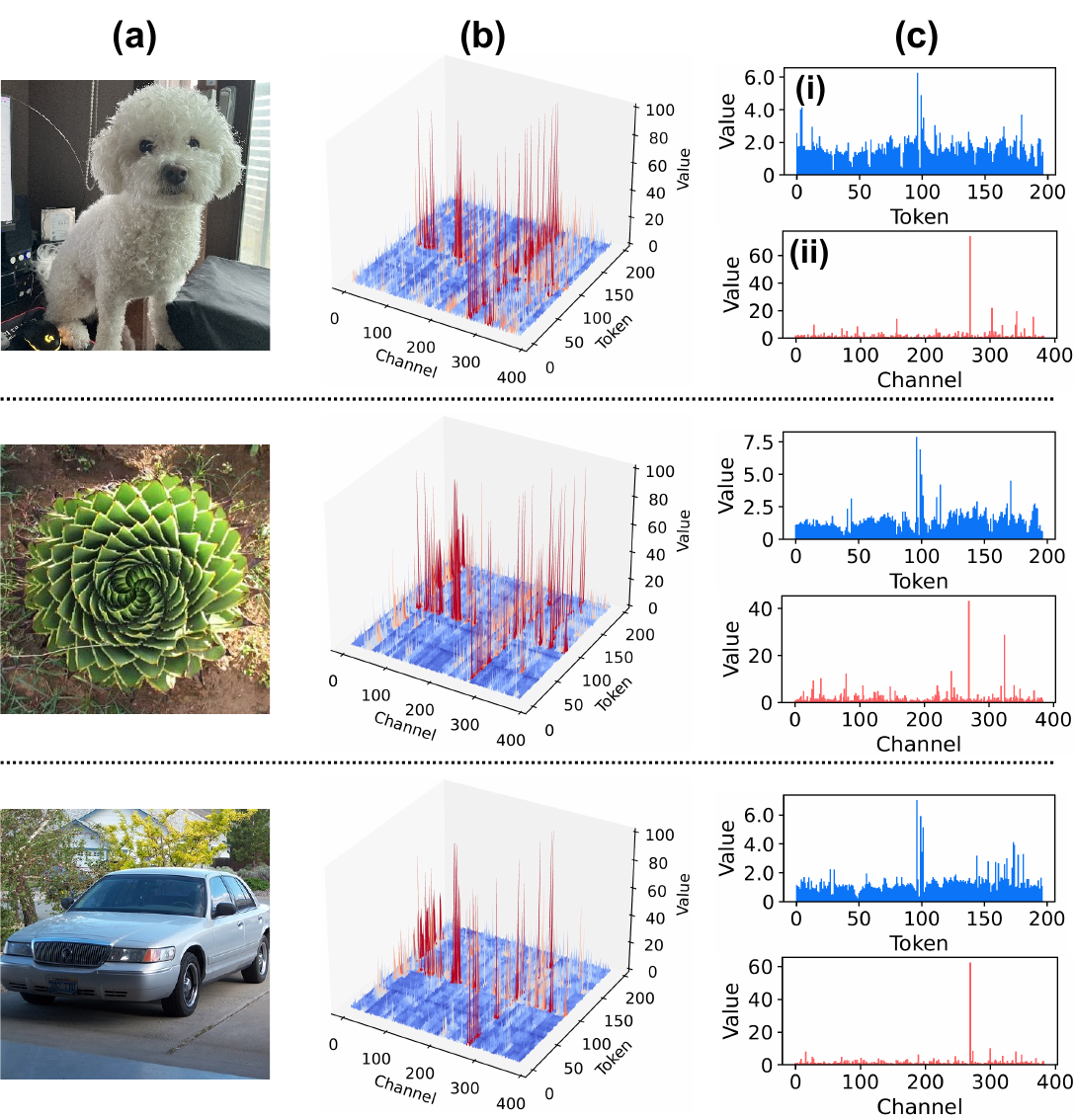}
\vspace{-6mm}
\caption{Distribution of the input activations of a 22nd out\_proj layer in Vim-Ti. (a) Images from 3 categories, (b) their corresponding activation distributions, and (c) the average values across (i) the channel dimension and (ii) the token dimension.}
\vspace{-4mm}
\label{fig:observation}
\end{figure}

\begin{figure*}
\centering
\includegraphics[width=0.95\linewidth]{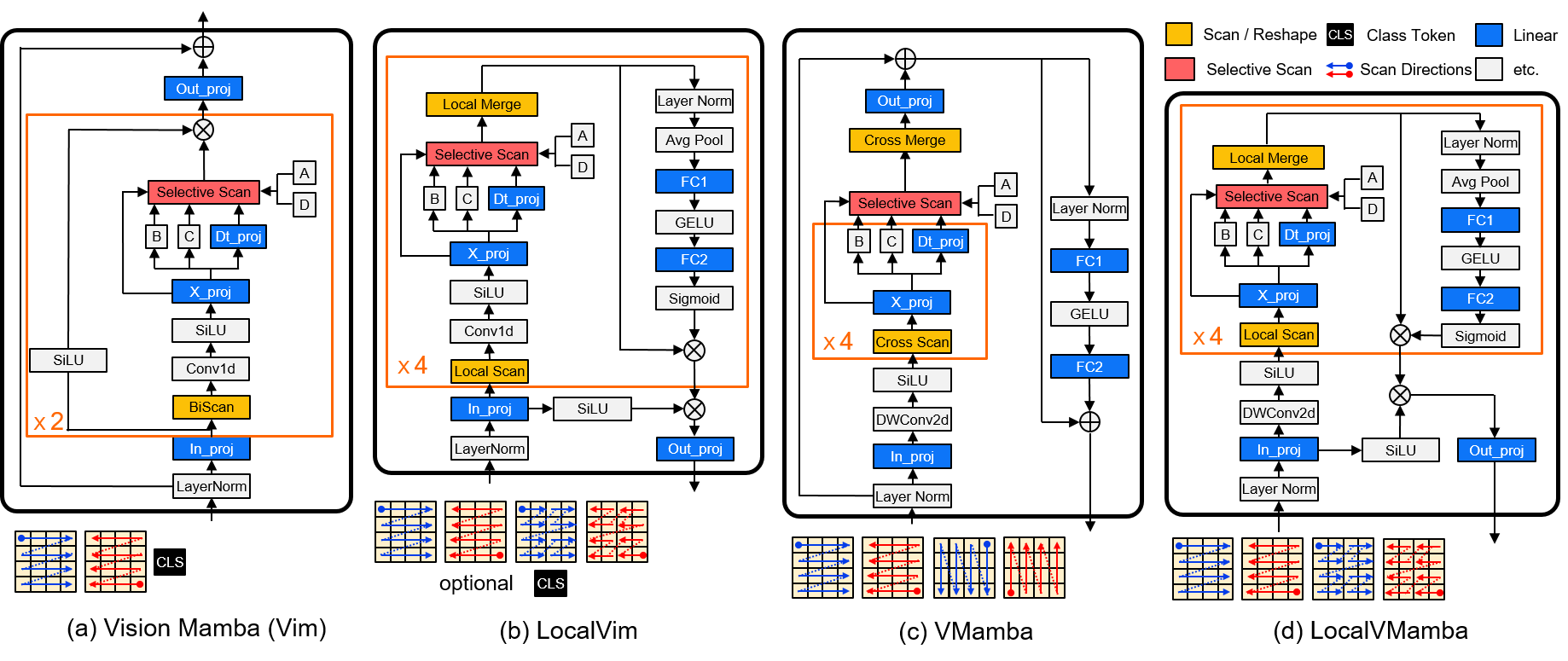}
\vspace{-3mm}
\caption{The Visual Mamba backbones consist of (a) Vision Mamba, (b) LocalVim, (c) VMamba, and (d) LocalVMamba. "x2" and "x4" indicate the repetition of operations based on scan directions. The square matrices beneath illustrate the scan method for each backbone.}
\vspace{-3mm}
\label{fig:structure}
\end{figure*}

\section{Introduction}

The State Space Model (SSM)~\cite{guefficiently, gu2021combining} was introduced to address the quadratic computational cost of transformers and to process sequential data more efficiently. An enhanced version of SSM, called Mamba~\cite{gu2023Mamba}, further improves this by updating the internal states selectively. Mamba has outperformed transformers and other sub-quadratic models across various language tasks~\cite{poli2023hyena, peng2023rwkv, fuhungry}, offering higher accuracy with relatively low computational cost. Recently, there has been growing interest in extending Mamba's capabilities to vision tasks~\cite{zhu2024vision, liu2024vMamba, huang2024localMamba, yang2024plainmamba}, often referred to as Visual Mamba. These efforts have introduced new module designs and features, such as class tokens for image data, demonstrating the superiority of Mamba in visual tasks.

In this study, we aim to enhance the cost-efficiency of Visual Mamba models through quantization. The performance advantage of the Visual Mamba can be further improved by reducing computational overhead and memory footprint via quantization. Our profiling results of existing Visual Mamba backbones revealed that a significant portion of execution time is dominated by linear operators (blue in \cref{fig:profile-kernel-fp}), which are well-suited for low-precision computations. This analysis suggests that Visual Mamba can sufficiently benefit from quantization in practice.

However, our observations identified a significant quality degradation when applying traditional post-training quantization (PTQ)\cite{banner2019post, nagel2021white, nagel2020up} techniques to Visual Mamba. Specifically, the structure of Visual Mamba (see \cref{fig:structure}), which sequentially processes image data tokens in a fixed order, results in an activation distribution that is particularly susceptible to quantization. The vulnerabilities of Visual Mamba can be categorized into three key items: 1) token-wise variance (\cref{fig:observation}c (i)), 2) channel-wise outliers (\cref{fig:observation}c (ii)), and 3) the long tail of activations (\cref{fig:observation_CLS}). As these challenges are not adequately addressed by conventional quantization methods, resolving them is crucial for maintaining output quality after quantization in Visual Mamba.

The weaknesses identified are prevalent across all Visual Mamba backbones proposed thus far, indicating that addressing them could yield broad benefits across a variety of models. Building on these insights, we introduce PTQ4VM, an effective and efficient post-training quantization (PTQ) scheme for Visual Mamba. To the our best knowledge, PTQ4VM is the first comprehensive study on quantization techniques for Visual Mamba. It is founded on two key techniques: Per-Token Static Quantization (PTS) and Joint Learning of Smoothing Scale and Step Size (JLSS). PTS is specifically designed to handle per-token variance, and we have carefully crafted it to be compatible with existing SmoothQuant method~\cite{xiao2023smoothquant}, which are effective in managing outliers within channels. Moreover, JLSS jointly optimizes the quantization parameters for PTS and scales for SmoothQuant, ensuring minimal discrepancies in output feature maps and preserving the network's functionality after quantization. Both PTS and JLSS are meticulously designed to maximize throughput to realize acceleration in practice, and we demonstrate the versatility and superiority of PTQ4VM through extensive experiments.

%------------------------------------------------------------------------
\section{Related Works}
\label{sec:formatting}
%-------------------------------------------------------------------------

%-------------------------------------------------------------------------
\subsection{Selective State Space model}
State Space Models (SSMs) \cite{gu2021combining, gu2022train} are linear time-invariant (LTI) systems that process sequential data through internal state variables. Originally designed for natural language processing (NLP) tasks, they often use an input-independent discretized formulation, expressed as follows:
\begin{align}
    \bar{A} &= \mathrm{exp}(\Delta A), \\
    \bar{B} &= (\Delta A)^{-1}(\exp(\Delta A) - I)\Delta B,
\end{align}
where $A\in \mathbb{R}^{N\times N}$, $B\in \mathbb{R}^{N\times 1}$, and $C \in \mathbb{R}^{1 \times N}$  are the parameters of the SSM, and 
$\Delta$ is the timescale parameter.
\begin{align}  
    h_t &= \bar{A}h_{t-1} + \bar{B}x_t,  \label{eq:ht} \\
    y_t &= Ch_t.
\end{align}
Recent research introduced Mamba \cite{gu2023Mamba}, a general language backbone that eliminates the linear time-invariant (LTI) property of SSMs, making them input-dependent. The discretized parameters for each input sequence of length $t$ are computed as follows, based on the input $x$:
\begin{align}
    B_t, \Delta_t, C_t &= \mathrm{Linear}_x(x_t), \\
    \bar{\Delta}_t &= \mathrm{softplus}\left(\mathrm{Linear}_\Delta(\Delta_t)\right), \\
    \bar{A}_t &=  \mathrm{exp}(\bar{\Delta}_t A), \\
    \bar{B}_t &= \bar{\Delta}_t B_t.
\end{align}
Mamba has gained significant attention for its ability to efficiently handle long sequence data with linear complexity. By introducing a selective scan that updates key information based on the input, it offers a distinct advantage over previous models that lacked this capability.

%-------------------------------------------------------------------------
\subsection{Visual Mamba Backbones}

Vision Mamba (Vim)~\cite{zhu2024vision} represents the first attempt to apply Mamba directly to vision tasks. Vim employs a CLS token, which is essential for its classification tasks. VMamba~\cite{liu2024vMamba}, on the other hand, features a backbone architecture that closely resembles the Swin Transformer~\cite{liu2021swin}. LocalMamba~\cite{huang2024localMamba} introduces two variants—LocalVim and LocalVMamba—based on the Vim and VMamba architectures, respectively, with enhancements in scan directions. Notably, among these LocalMamba variants, the LocalVim-T$^\dagger$ model makes use of the CLS token. \cref{fig:structure} illustrates the detailed modular design of each backbone.

%-------------------------------------------------------------------------

\subsection{Post-training Quantization}
\label{sec:ptq}

Quantization is currently the most commercially successful optimization method for leveraging the benefits of low-precision representations. Early research primarily focused on quantization-aware training (QAT)~\cite{esser2019learned, zhou2016dorefa, Lee_2023_ICCV, shin2023nipq}, but its popularity has waned due to the high cost of the training process. Consequently, post-training quantization (PTQ)~\cite{librecq, nagel2019data, nagel2020up, weiqdrop, so2024temporal, kim2024hlq} has emerged as a key interest.

In this work, we aim to develop a PTQ scheme for linear quantization, focusing on accelerating computation through integer arithmetic while minimizing transformation costs. We adopt the conventional PTQ approach~\cite{jacob2018quantization}, utilizing per-channel symmetric quantization for weights and per-tensor asymmetric for activations. Given an input activation $X$, the quantized value $\hat{X}$ is calculated as follows:
\begin{align}
    \Delta_X &= \frac{max(X) - min(X)}{2^{b} - 1}, \\
    \epsilon_X &= \frac{-min(X)}{\Delta_X}, \\
    % \hat{X} &= \Delta_X\cdot (clip(\lceil \frac{X}{\Delta_X} \rfloor + \epsilon_X, 0, 2^b-1) + \epsilon_X).
    \hat{X} &= \Delta_X\cdot \left(clip(\lceil \frac{X}{\Delta_X} \rfloor + \epsilon_X, 0, 2^b-1) - \epsilon_X\right).
\end{align}
where $b$ represents the number of bits, $\Delta_X$ denotes the quantization step for the activation, and $\epsilon_X$ represents the quantization offset.

In the case of the given weight $W$, the quantized value $\hat{W}$ using symmetric quantization is calculated as follows:
\begin{align}
    \Delta_W &= \frac{max(abs(W))}{2^{b-1} - 1}, \\
    \hat{W} &= \Delta_W\cdot  clip(\lceil \frac{W}{\Delta_W} \rfloor, -2^{b-1}+1, 2^{b-1}-1).
\end{align}
where $\Delta_W$ represents the quantization step for the weight, and $\epsilon_W$ is omitted in symmetric quantization.

For the baseline, the quantization range is determined using min-max values~\cite{nagel2021white}. While this method has been empirically proven to perform well in CNN-based networks, it is less effective for Visual Mamba, which exhibits significantly different characteristics.

%-------------------------------------------------------------------------

\subsection{SmoothQuant}
\label{sec:smoothquant}
Quantization has also been actively studied in the context of Large Language Models (LLMs)~\cite{brown2020language, radford2019language, touvron2023llama, zhang2022opt}. LLMs present unique challenges that complicate the quantization process, such as the presence of channel-wise outlier activations~\cite{dettmers2022gpt3, lee2024owq, lee2024qeft}. These outliers cause a significant increase in quantization error by enlarging the quantization step size. To address this, SmoothQuant~\cite{xiao2023smoothquant} was proposed, aiming to mitigate the impact of outliers by shifting the quantization complexity from activations to weights, all without introducing additional computational overhead. In SmoothQuant, given the input activation $X$ and weight $W$, the normalization scale is calculated as:
\begin{equation}
\label{eq:smooth_scale}
s = \sqrt{\frac{max|X|}{max|W|}} \in \mathbb{R}^{D_{in}},
\end{equation} 
where $D_{in}$ is the input channel. 
The computed scale is then applied to both $X$ and $W$ respectively, adjusting each tensor to ensure that the overall output remains consistent:
\begin{equation}
Y = (Wdiag(s)) \cdot (diag(s)^{-1}X).
\label{eq:smooth}
\end{equation} 
This method mitigates activation outliers, thereby reducing quantization errors. According to our observation, in Visual Mamba, although the underlying causes may differ, we observed a similar phenomenon where outliers occur only in specific activation channels. The details of observation and the solution will be provided at \Cref{sec:challenges}.

%------------------------------------------------------------------------
\begin{figure}
\centering
\includegraphics[width=0.95\linewidth]{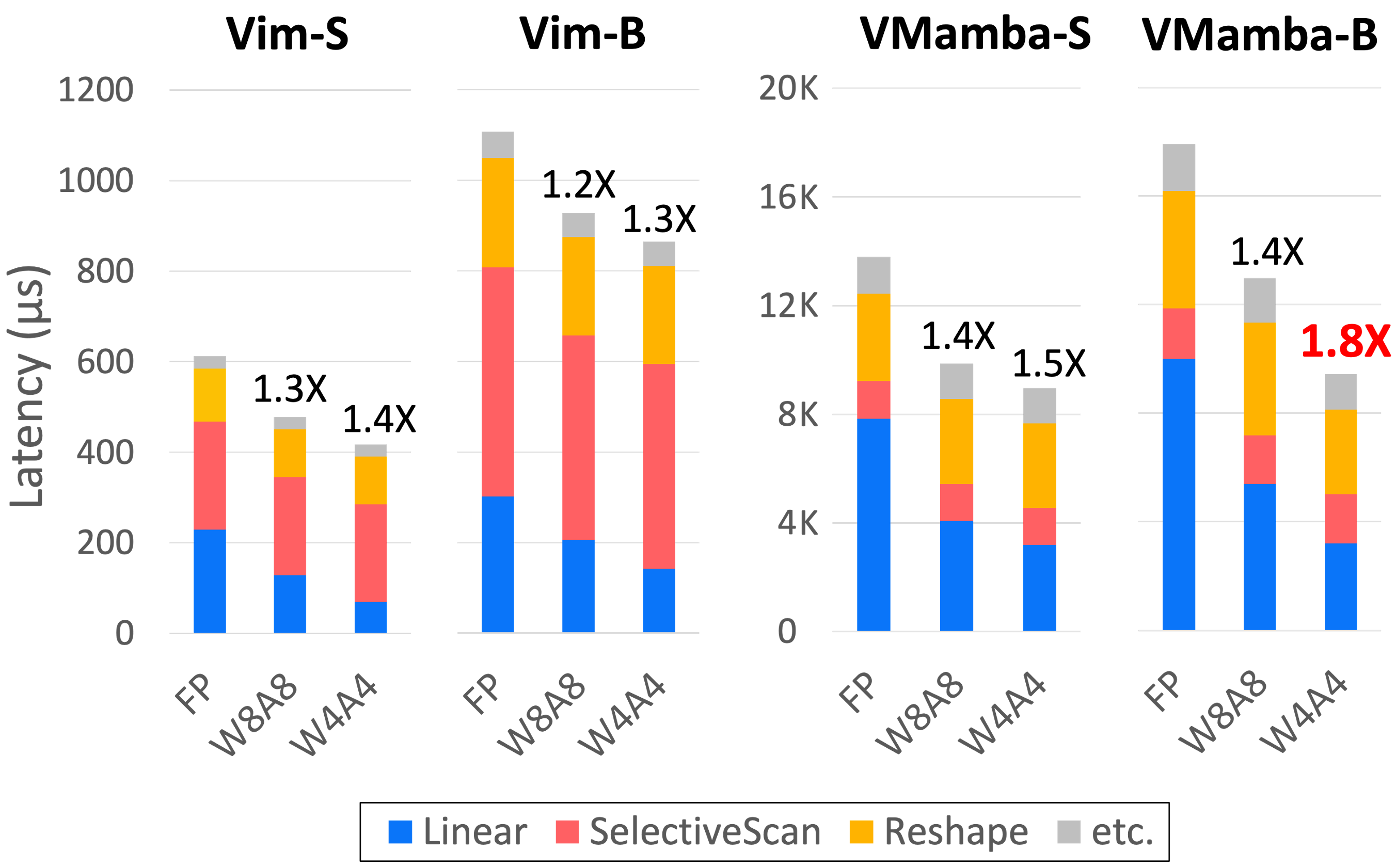}
\vspace{-3mm}
\caption{Profiling results of Visual Mamba backbones on an RTX 3090. The numbers above the bars indicate the speedup.}
\vspace{-4mm}
\label{fig:profile-kernel-fp}
\end{figure}

%------------------------------------------------------------------------
\section{Analysis for Quantization Target}
Before applying quantization, we analyzed whether existing Visual Mamba backbones benefit from quantization. We profiled the two models among models presented in \cref{fig:structure} on the GPU and analyzed which components would be most suitable for quantization. 

As shown in \cref{fig:profile-kernel-fp}, the operations that consistently consume the most time across all models are the linear layer (blue section in \cref{fig:profile-kernel-fp}), selective scan~\cite{gu2023Mamba} (red section), and reshape operation (yellow section). Our analysis of the quantization benefits for these three operations leads to the following conclusions: 1) For the linear operation, significant performance improvements can be achieved by applying quantization using INT8/INT4 operators~\cite{jacob2018quantization}. In the case of the selective scan, it has been optimized to minimize memory bottlenecks through the use of registers and shared memory, so it is difficult to expect performance gains through quantization. More concerning, our experimental results, reported in \Cref{tab:hidden_state_quant} left, show that despite applying quantization only to the hidden state $h(t)$ in \cref{eq:ht}, it is far more vulnerable than anticipated,  outweighing the expected performance gains. The reshape operation is used to change the data layout between adjacent operators for optimal speed or to rearrange data order for scan operations. However, the overhead caused by reshape operations is difficult to be mitigated by quantization.

Based on this analysis, we concluded that focusing on quantizing the linear layers is a reasonable approach. As illustrated in \cref{fig:profile-kernel-fp}, applying PTQ4VM to the linear layers can reduce latency by up to 1.83$\times$ on the real GPUs, making it a highly appealing option.

\section{Challenges of Linear Layer Quantization}
\label{sec:challenges}
Based on the previous analysis, we identified the linear operators in Visual Mamba as key targets for optimization. However, despite this focused optimization effort, Visual Mamba still shows a significant drop in quality, regardless of the backbone used. Further investigation revealed that its sequential access to visual tokens for information accumulation causes abnormal activation distributions, making quantization especially difficult. We categorized these difficulties into three main items: token-wise variance, channel-wise outliers, and the long tail of activations. In the following section, we will provide a more detailed explanation of the problems we identified.

\begin{table}[t]
  \centering
  \small
  \setlength\tabcolsep{3pt}
  \begin{tabular}{cc c||c cc}
    \toprule
    \textbf{Quant Target} & \textbf{Top-1 Acc.} && & \textbf{Method} & \textbf{Top-1 Acc.} \\
    %\midrule
    \cline{1-2} \cline{5-6}
    FP16        & 76.1  && & FP16 & 76.1 \\
    $h(t)$ only & 6.8 && & INT8 Baseline & 57.8 \\
    Linear only & 57.8 && & + CLS Tok. FP16 & 73.6 \\
    \bottomrule
  \end{tabular}
  \vspace{-2mm}
  \caption{INT8 accuracy results (\%) on Vim-Ti (Left) across different quantization targets, (Right) with FP16 CLS token.}
  \vspace{-4mm}
  \label{tab:hidden_state_quant}
\end{table}

\begin{figure}
\centering
\includegraphics[width=\linewidth]{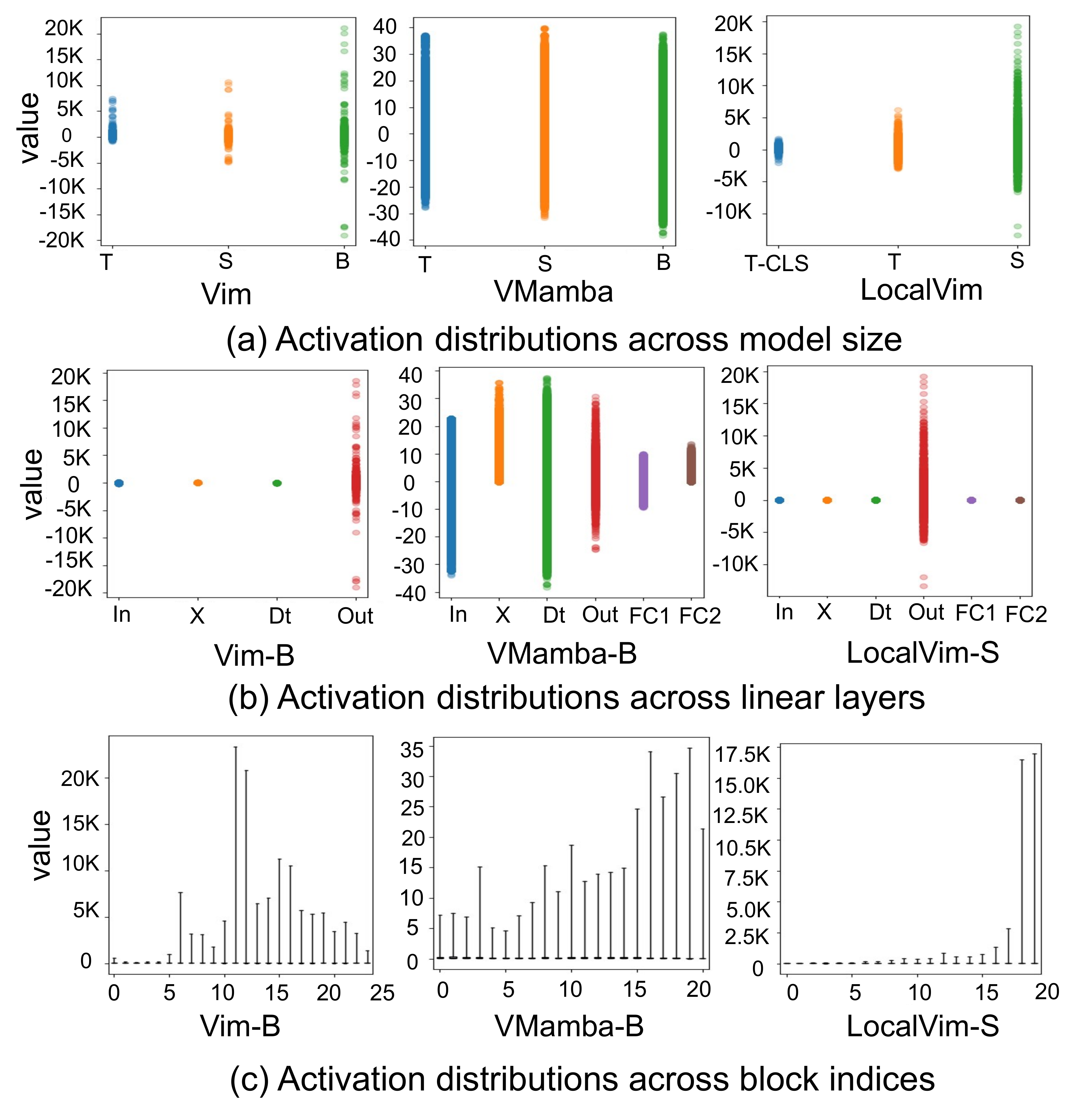}
\vspace{-8mm}
\caption{Observation on activation distributions across (a) model size, (b) types of linear layers, and (c) block indices. (c) shows the activation distribution of the out\_proj layer.}
\vspace{-5mm}
\label{fig:distribution_tendency}
\end{figure}

\subsection{Observation 1: Token-wise Variance}
\label{sec:observation1}
We began by analyzing the activation distribution by feeding various images into Visual Mamba. \cref{fig:observation} illustrates three representative cases. The results show that specific token positions, such as position 97, consistently displayed similar activation patterns, regardless of the image class or features. This behavior appears to stem from Visual Mamba's unique architecture, where image patches are processed sequentially in a predetermined order, as suggested by the scan directions in \cref{fig:structure}. Additionally, we observed that token-wise activation variance increases significantly in the middle to later blocks of the network compared to the early blocks, with the differences between tokens also becoming more pronounced (\cref{fig:distribution_tendency}c). This suggests that the cumulative effect over the blocks amplifies the imbalance, making larger networks more challenging to quantize.

On the other hand, the Visual Mamba backbone can be categorized into two types of implementations: those incorporating the CLS token and those relying solely on visual tokens. In the case of the Vim model, which uses the CLS token, it is noticeable that the magnitude of the CLS token is significantly smaller than that of the visual tokens, regardless of the input. This presents a challenge, as the CLS token is crucial for downstream tasks like classification, making it particularly vulnerable to quantization errors. As shown in \Cref{tab:hidden_state_quant} right, preserving the CLS token in FP16 format substantially recovers most of the accuracy loss compared to fully quantizing the model’s linear layers to INT8. For networks that use the CLS token, we should reduce the quantization error for this special token.

Due to the variation in token-wise activation, conventional tensor-wise quantization (as shown in \cref{fig:overall}b) is inevitably suboptimal for individual tokens. This becomes a key factor in increasing quantization error and highlights the need for an appropriate solution. One potential approach is per-token dynamic quantization, which adjusts the quantization range for each token based on its distribution during input processing. However, this method has a significant drawback: activation statistics must be computed online for each input, slowing down inference as shown in \Cref{tab:Latency comparison}.

To fully preserve the benefits of low-precision arithmetic in the Visual Mamba backbone as well as maintain the quality of output, it is essential to explore methods that address token-wise variance to minimize the accuracy gap with per-token dynamic quantization while maintaining a speed comparable to per-tensor static quantization.

\subsection{Observation 2: Channel-wise Outliers}
\label{sec:observation2}
The second observation is that activation outliers tend to occur in a few specific input channels, regardless of the input (\cref{fig:observation}c (ii)). This phenomenon is common across all backbones, where a small number of activation outliers lead to larger quantization steps, causing information loss in other tokens. It's important to note that this issue stems from a different dimension than the one discussed in the first observation. Even when per-token dynamic quantization, a potential solution for the previous issue, is applied, the same step size must be used for all channels within a token. As a result, it fails to mitigate the performance degradation caused by these activation outliers.

As discussed in \Cref{sec:smoothquant}, although the underlying causes may differ, activation outliers have also been observed in Large Language Models (LLMs) with transformer architectures. When we applied smoothing to Visual Mamba backbones, we noted quality improvements at 8-bit and 6-bit quantization (see \Cref{tab:classification_results}). This suggests that smoothing should also be integrated into the quantization process for Visual Mamba. Accordingly, our proposed PTQ4VM is specifically designed to incorporate it.

\begin{figure}
\centering
\includegraphics[width=\linewidth]{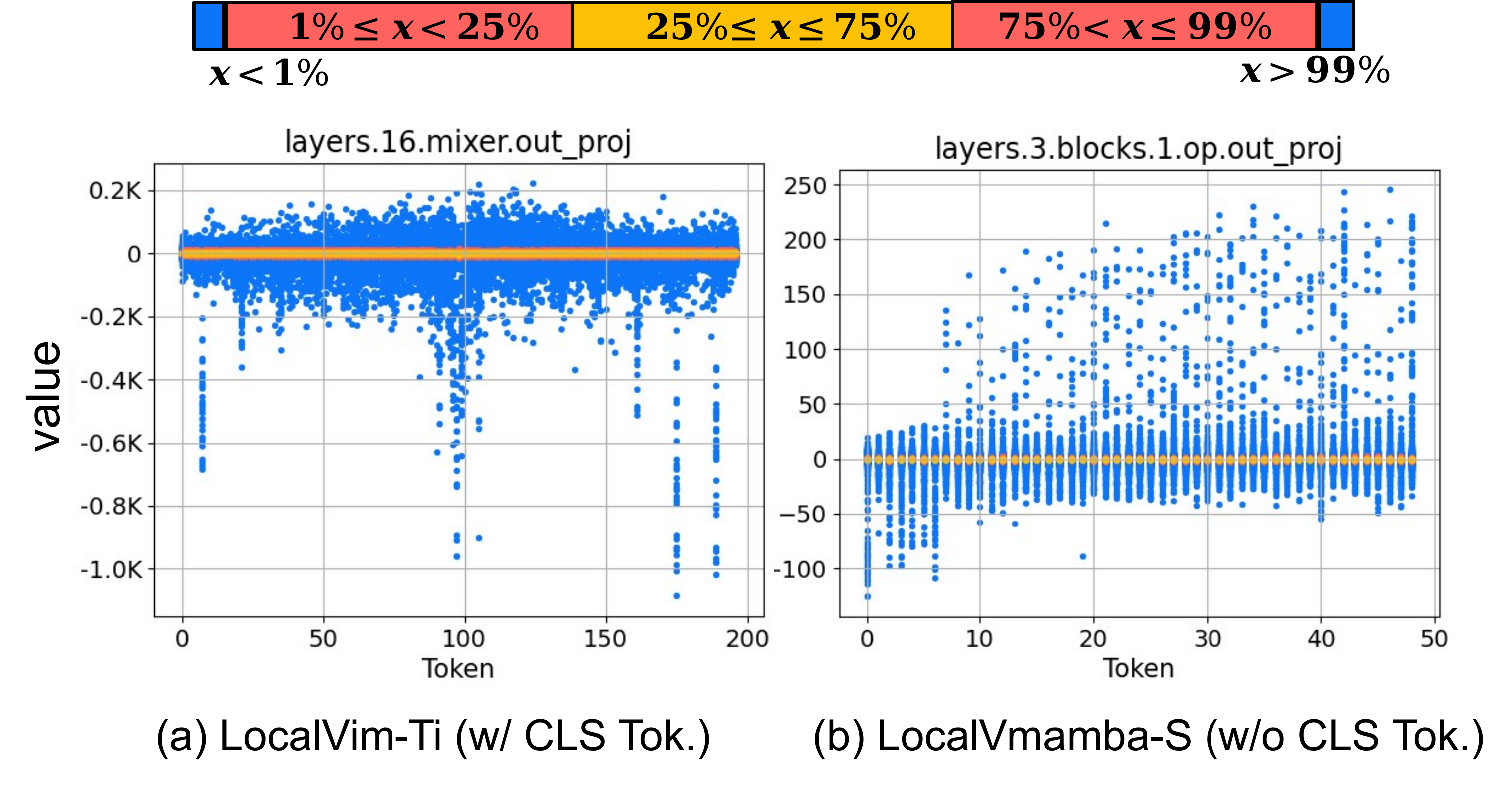}
\vspace{-8mm}
\caption{Comparison of activation distribution on LocalMamba backbone depending on whether CLS Token is utilized or not.}
\vspace{-4mm}
\label{fig:observation_CLS}
\end{figure}

\subsection{Observation 3: Long tail of Activation}
\label{sec:observation3}
The activation distribution of Visual Mamba exhibits a distinctive characteristic. As illustrated in \cref{fig:observation_CLS}, approximately 98\% of the activation values are concentrated within a very narrow range (represented by the red and yellow areas in the plot). The data spread across a wider range corresponds to the top and bottom 1\% of values, indicating a long-tailed distribution. Similar to the outliers seen in Observation 2, this long tail extends the quantization range too far, leading to a loss of information from tokens near zero. Notably, as the backbones in the Vision Mamba series feature tails extending beyond 10,000, smoothing outliers with SmoothQuant alone is insufficient to fully resolve the issue.

Previous studies~\cite{bhalgat2020lsq+, li2021brecq} have emphasized the importance of introducing truncation to balance the trade-off between clipping and rounding errors. Given the long tail of activation, truncation is also crucial for Visual Mamba. However, similar to dynamic quantization, input-dependent truncation leads to significant performance degradation due to the associated online costs. To minimize quantization errors while preserving the acceleration benefits, it is essential to determine the optimal static truncation range that is generally applicable across elements within the same token position. 

\section{PTQ4VM}

\begin{figure*}
\centering
\includegraphics[width=\linewidth]{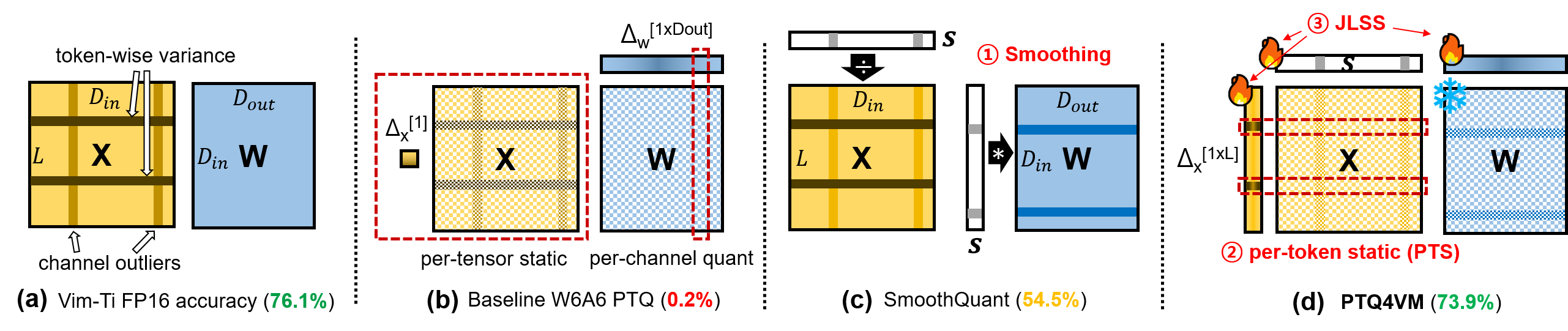}
\vspace{-8mm}
\caption{Comparisons of quantization methods. (b) and (c) uses per-tensor static quantization for activation.}
\vspace{-5mm}
\label{fig:overall}
\end{figure*}

In this section, we propose PTQ4VM (\cref{fig:overall}d), a post-training quantization method designed to effectively address the three challenges identified in~\Cref{sec:challenges}. 
It mainly consists of two parts: First, a Per-Token Static (PTS) quantization to handle Observation 1 and 2 (\Cref{sec:PTS}). Next, the JLSS method to find the optimal smooth scale and step size to address Observation 3 (\Cref{sec:JLSS}).

\subsection{Per-Token Static (PTS) Quantization}
\label{sec:PTS}
   We propose Per-Token Static (PTS) quantization, a simple yet powerful method to address Observations 1 and 2 with minimal computational overhead. Since the token length in Visual Mamba is predetermined by the fixed input size, we can allocate the quantization step size and zero offset for each token of length \(L\) using a calibration dataset. The weight step size \(\Delta_W \in \mathbb{R}^{D_{out} \times 1}\) and activation step size \(\Delta_X \in \mathbb{R}^{1 \times L}\) are vectorized scaling factors determined by the tensor dimensions. Specifically, \(\Delta_W\) represents the per-output channel scaling factors, while \(\Delta_X\) denotes the per-token scaling factors along the activation sequence. These scaling factors are used to adjust the range of the integer-mapped tensors during the quantization process.
    
    After quantization, the integer-mapped weight \(\bar{W} \in \mathbb{R}^{D_{out} \times D_{in}}\) and activation \(\bar{X} \in \mathbb{R}^{D_{in} \times L}\) can be efficiently multiplied via integer operations. To incorporate the scaling factors, the final output is computed through a sequence of matrix multiplications involving diagonal matrices formed from \(\Delta_W\) and \(\Delta_X\). Here, \(\text{diag}(\Delta_W)\) and \(\text{diag}(\Delta_X)\) represent diagonal matrices where the elements of \(\Delta_W\) and \(\Delta_X\) are placed along the main diagonal, and the operator \(\cdot\) denotes standard matrix multiplication. The final output $Y$ is expressed as follows:
\begin{align}
    Y &\approx (\text{diag}(\Delta_W)\cdot\bar{W})\cdot(\bar{X}\cdot\text{diag}(\Delta_X)) \notag \\
      &= \text{diag}(\Delta_W)\cdot (\bar{W} \cdot \bar{X}) \cdot \text{diag}(\Delta_X).
\end{align}
    
    Note that we omitted the zero offset (\(\epsilon_X\)) for simplicity of explanation. If \(\Delta_W\) and \(\Delta_X\) can be fused across adjacent linear layers, the output can be computed using only low-precision operations without the need for additional element-wise scaling. PTS predetermines the step size statically, resulting in significantly lower overhead compared to per-token dynamic methods, thereby gaining an advantage in acceleration. As reported in \Cref{tab:Latency comparison}, our method demonstrates a 1.3$\times$ speed improvement over per-token dynamic approaches. Furthermore, PTS is a modification applied orthogonally to SmoothQuant, offering the advantage of compatibility with smoothing techniques. Through this combined approach, we can address the aforementioned problems from observation 1 and 2.

\subsection{Joint Learning of Smoothing Scale and Step Size (JLSS)}
\label{sec:JLSS}
To address the third challenge, the long tail of activations, we introduce Joint Learning of Smoothing Scale and Step Size (JLSS). 
The core objective of JLSS is to identify the optimal values for the smoothing scale \( s\) and step sizes \( \Delta_X \) and \( \Delta_W \), ensuring minimal deviations in the output tensor. To achieve this, JLSS employs a three-stage method for concurrent optimization of these values.

In the first stage, we apply smoothing across all linear layers using a small calibration set to reduce outliers. The smoothing scale is initialized as described in \cref{eq:smooth}. 
In the second stage, through a grid search, we initialize the $\Delta_X$ and $\Delta_W$ values to minimize the L2 loss for the calibration set.
The loss for searching $\Delta_X$ is defined as the L2 distance between the FP16 values of $X$ and the quantized $\hat{X}$ for each layer, with the loss for $\Delta_W$ defined similarly.
Finally, in the third stage, we sequentially tune $s$, $\Delta_X$, and $\Delta_W$ to minimize the quantization error, starting from the earlier blocks and updating the subsequent ones, using gradient descent. Specifically, the quantization error is defined as the cosine similarity between the FP16 and quantized block outputs. Throughout this process, the quantized integer weights $\bar{W}$ remain fixed, while only the $s$, $\Delta_X$, and $\Delta_W$ are learned. This approach initializes each value close to optimal in a layer-wise manner and then updates only those parameters with gradient descent in a block-wise manner for a few steps. PTQ4VM generates any quantized model within 15 minutes on a single RTX 3090 GPU.

The optimally learned quantization parameters play a key role in minimizing quantization errors and preserving network performance, even in the presence of outliers and long-tailed distributions. After training, these parameters facilitate acceleration using low-precision arithmetic, maximizing performance gains on actual GPU hardware.

\section{Experiments}
In order to demonstrate the superiority of the PTQ4VM, we conducted comprehensive experiments across various computer vision tasks, including image classification, object detection, and instance segmentation. We employed MinMax quantization, as described in \Cref{sec:ptq}, and SmoothQuant, detailed in \Cref{sec:smoothquant}, as our baselines. Both weights and activations of linear layers were quantized, with the notation W8A8 representing 8-bit weights and 8-bit activations, respectively.

\subsection{Quantization Setting}
For all tasks, we used the same calibration sets to minimize bias. For the Image Classification task, we randomly sampled 256 images from the ImageNet-1K \cite{russakovsky2015imagenet} training set. For the Object Detection and Instance Segmentation tasks, we randomly sampled 16 images from the MSCOCO-2017 \cite{lin2014microsoft} training set as the calibration set. Detailed experimental configurations are provided in the supplementary.

\begin{table}[t]
    \centering
    \small
    \setlength\tabcolsep{3pt}
    \resizebox{\columnwidth}{!}{
    \begin{tabular}{cccccc}
    \toprule
    \multirow{2}{*}{\textbf{Model}} & \multirow{2}{*}{\textbf{Method}} & \multicolumn{4}{c}{\textbf{Top-1 Accuracy (\%)}} \\
                              &  & \textbf{FP16} & \textbf{W8A8} & \textbf{W6A6} & \textbf{W4A4} \\
    \midrule
    \multirow{3}{*}{Vim-Ti}
          & MinMax & \multirow{3}{*}{76.1} & 57.8 & 1.7 & 0.1 \\
          & SmoothQuant &  & 74.7 & 54.5 & 0.1 \\
          & PTQ4VM (ours) &  & \textbf{75.8} & \textbf{73.9} & \textbf{56.4} \\
    \midrule
    \multirow{3}{*}{Vim-S}
          & MinMax & \multirow{3}{*}{80.5} & 79.4 & 27.3 & 0.1 \\
          & SmoothQuant &  & 80.1 & 73.7 & 0.2 \\
          & PTQ4VM (ours) &  & \textbf{80.5} & \textbf{79.7} & \textbf{69.6} \\
    \midrule
    \multirow{3}{*}{Vim-B}
          & MinMax & \multirow{3}{*}{81.9} & 75.8 & 0.5 & 0.1 \\
          & SmoothQuant &  & 79.9 & 52.3 & 0.1 \\
          & PTQ4VM (ours) &  & \textbf{80.3} & \textbf{79.7} & \textbf{55.6} \\
    \midrule
    \multirow{3}{*}{VMamba-T}
          & MinMax & \multirow{3}{*}{82.6} & 82.6 & 81.6 & 1.2 \\
          & SmoothQuant &  & 82.6 & 81.8 & 1.7 \\
          & PTQ4VM (ours) &  & \textbf{82.6}& \textbf{82.4} & \textbf{80.6} \\
    \midrule
    \multirow{3}{*}{VMamba-S}
          & MinMax & \multirow{3}{*}{83.6} & 83.6 & 82.8 & 1.1 \\
          & SmoothQuant &  & 83.6 & \textbf{83.6} & 4.5 \\
          & PTQ4VM (ours) &  & \textbf{83.6} & 83.5 & \textbf{82.3} \\
    \midrule
    \multirow{3}{*}{VMamba-B}
          & MinMax & \multirow{3}{*}{83.9} & 83.6 & 73.9 & 0.3 \\
          & SmoothQuant &  & 83.8 & 83.3 & 1.2 \\
          & PTQ4VM (ours) &  & \textbf{83.9} & \textbf{83.8} & \textbf{82.8} \\
    \midrule     
    \multirow{3}{*}{LocalVim-T$^\dagger$}
          & MinMax & \multirow{3}{*}{78.1} & 76.2 & 42.4 & 0.1 \\
          & SmoothQuant &  & 76.7 & 52.6 & 0.2 \\
          & PTQ4VM (ours) &  & \textbf{77.6} & \textbf{76.1} & \textbf{53.7} \\
    \midrule
    \multirow{3}{*}{LocalVim-T}
          & MinMax & \multirow{3}{*}{76.2} & 75.6 & 62.2 & 0.4 \\
          & SmoothQuant &  & 75.9 & 65.5 & 0.7 \\
          & PTQ4VM (ours) &  & \textbf{76.2} & \textbf{75.7} & \textbf{67.2} \\
    \midrule
    \multirow{3}{*}{LocalVim-S}
          & MinMax & \multirow{3}{*}{81.1} & 80.9 & 63.5 & 0.2 \\
          & SmoothQuant &  & 81.0 & 69.7 & 0.6 \\
          & PTQ4VM (ours) &  & \textbf{81.1} & \textbf{80.3} & \textbf{68.4} \\
    \midrule
    \multirow{3}{*}{LocalVMamba-S}
          & MinMax & \multirow{3}{*}{83.7} & 83.5 & 80.8 & 2.5 \\
          & SmoothQuant &  &83.6 & 81.9 & 12.0\\
          & PTQ4VM (ours) &  & \textbf{83.7} & \textbf{83.4} & \textbf{81.2} \\
    \bottomrule
    \end{tabular}
    }
    %\end{adjustbox}
    \vspace{-2mm}
    \caption{ImageNet Top-1 validation accuracy comparison of quantization methods on various models. LocalVim$^\dagger$ indicates a model that uses the CLS token.}
    \vspace{-4mm}
    \label{tab:classification_results}
\end{table}

\subsection{Image Classification}

 To validate the universal applicability of our proposed method across various Visual Mamba backbones, we compared PTQ accuracy for the four architectures presented in \cref{fig:structure}. \Cref{tab:classification_results} shows the top-1 accuracy results on the ImageNet-1K validation set. PTQ4VM consistently achieved significantly higher accuracy compared to other methods across all backbones and quantization options. 

The experimental results for the Vim family with CLS tokens reveal significant accuracy degradation when using MinMax, even at W8A8, as it fails to account for token-wise variance. Similarly, SmoothQuant struggles to retain essential CLS token information due to the same issue, leading to a 29.6\% accuracy loss at 6-bit for Vim-B. In contrast, PTQ4VM addresses all the challenges posed by Visual Mamba, demonstrating strong performance with less than 0.5\% accuracy loss at 8-bit for Vim-Ti/S and LocalVim-T$^\dagger$, while maintaining acceptable quality at 4-bit, where other methods fall short. Despite Vim-B having higher FP16 accuracy than Vim-S, its quantized performance lags across all methods, primarily due to outliers exceeding magnitudes of 20K. These findings suggest that Visual Mamba models with excessively large activation magnitudes may be inherently vulnerable to compression techniques.

In the VMamba family, all methods show lossless quality at W8A8 compared to FP16, largely due to its significantly smaller activation ranges compared to other model families. This unique characteristic can be linked to the absence of gating functions, as seen in \cref{fig:structure}b, which differentiates it from other architectures. However, while other methods struggle with W4A4, only PTQ4VM manages to maintain quality on VMamba-S/B with less than 1.3\% degradation.

\subsection{Object Detection and Instance Segmentation}

\begin{table}[t]
\centering
\small
\begin{tabular}{c|cccc}
\toprule
\multirow{2}{*}{\textbf{Backbone}} & \multirow{2}{*}{\textbf{Method}} & \multirow{2}{*}{\textbf{Bit}} & \multicolumn{2}{c}{\textbf{AP}} \\ \cline{4-5} 
 &  &  & \textbf{AP$^{b}$} & \textbf{AP$^{m}$} \\ \hline
\multirow{10}{*}{VMamba-T} & - & FP16 & 47.0 & 42.3 \\ \cline{2-5}
 & MinMax & W8A8 & 46.9 & 42.2 \\ 
 & SmoothQuant & W8A8 & 46.9 & 42.2 \\ 
 & PTQ4VM (ours) & W8A8 & \textbf{47.0} & \textbf{42.3} \\  \cline{2-5}
 & MinMax & W6A6 & 46.2 & 41.5 \\ 
 & SmoothQuant & W6A6 & 45.5 & 40.9 \\ 
 & PTQ4VM (ours) & W6A6 & \textbf{46.7} & \textbf{42.1} \\ \cline{2-5}
 & MinMax & W4A4 & 0.3 & 0.3 \\ 
 & SmoothQuant & W4A4 & 0.5 & 0.4 \\ 
 & PTQ4VM (ours) & W4A4 & \textbf{43.5} & \textbf{39.4} \\ 
\bottomrule
\end{tabular}
\vspace{-2mm}
\caption{Results of object detection and instance segmentation.}
\vspace{-2mm}
\label{tab:Detection_result}
\end{table}

\begin{figure}
\centering
\includegraphics[width=0.9\linewidth]{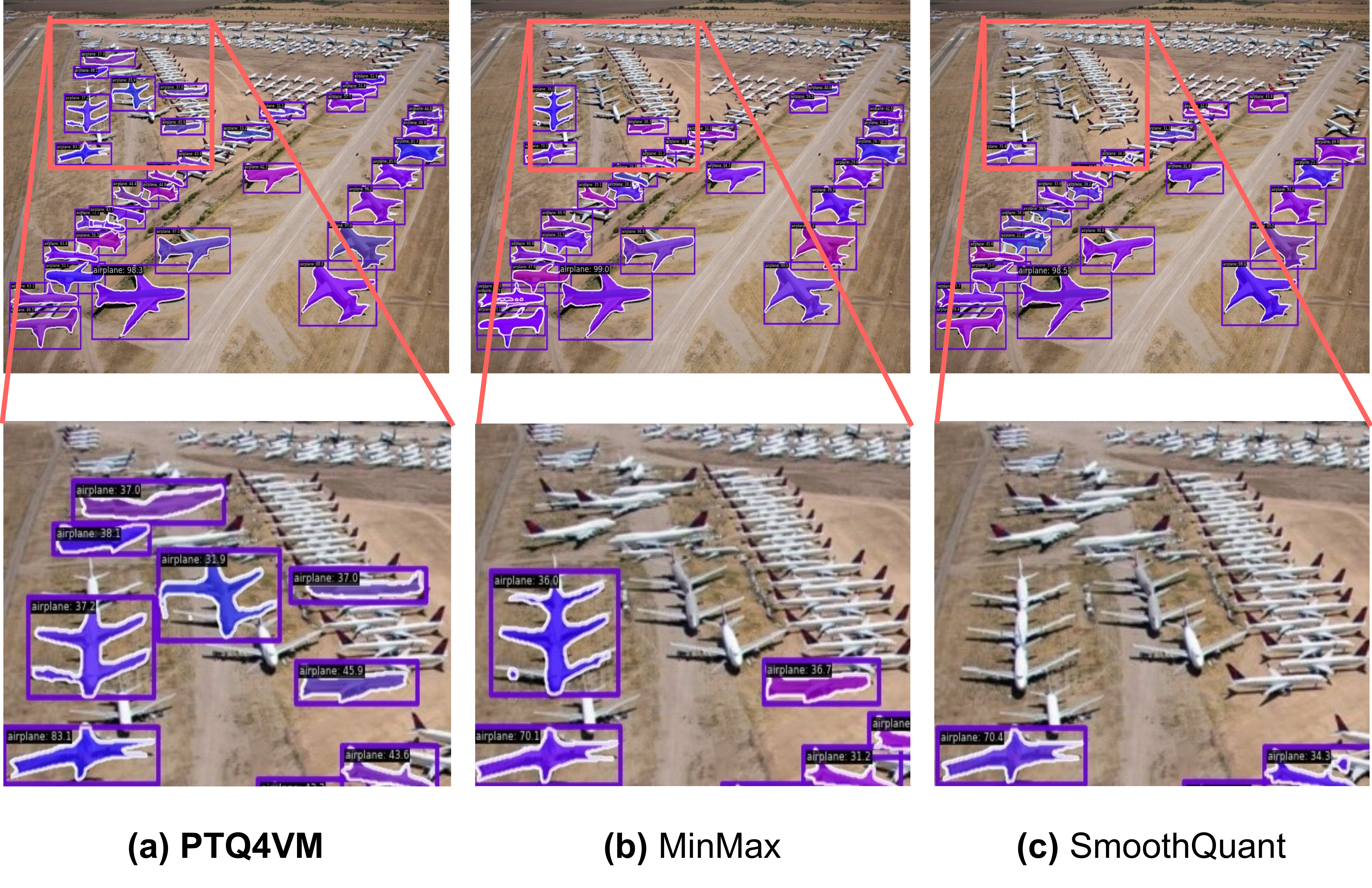}
\vspace{-2mm}
\caption{Qualitative results for the object detection and instance segmentation task. Applied W6A6 quantization to VMamba-T.}
\vspace{-4mm}
\label{fig:Detection figure}

\end{figure}

To demonstrate that PTQ4VM performs in downstream tasks, we evaluated its performance in Object Detection and Instance Segmentation tasks. We used the official checkpoint trained with Mask R-CNN\cite{he2017mask} 3X MS schedule using VMamba-T backbone for our experiments, evaluating on images cropped to 1280$\times$800. As reported in \Cref{tab:Detection_result}, while MinMax and SmoothQuant both fail at W4A4, PTQ4VM demonstrates superior performance with less than 3.5\% score degradation. At W6A6, SmoothQuant shows lower scores than MinMax, which can be attributed to the degradation caused by the difficulty imposed on weights by smoothing. In contrast, PTQ4VM achieves near-lossless performance with less than 0.3\% AP$^{b}$ and 0.2\% AP$^{m}$ degradation, due to optimal tuning by JLSS. Notably, the qualitative results in \cref{fig:Detection figure} validate that PTQ4VM maintains the network's quality well, preserving even small details.

\begin{table}[t]
\small
\centering
\setlength\tabcolsep{3pt}
%\resizebox{\columnwidth}{!}{
\begin{tabular}{@{}ccccc@{}}
\toprule
\multirow{2}{*}{\textbf{Model}}    & \multirow{2}{*}{\textbf{Method}} & \multicolumn{2}{c}{\textbf{Latency (ms)}}      & \multirow{2}{*}{\textbf{Speedup}} \\ \cmidrule(lr){3-4}
                          &                         & \textbf{FP16}                    & \textbf{W4A4}   &                        \\ \midrule
\multirow{3}{*}{VMamba-T} & Per-tensor static             & \multirow{3}{*}{60.33}  & 33.87  & 1.78$\times$                   \\
                          & Per-token dynamic       &                         & 44.13  & 1.37$\times$                   \\
                          & PTS (ours)              &                         & 34.00  & 1.77$\times$                   \\ \midrule
\multirow{3}{*}{VMamba-S} & Per-tensor static            & \multirow{3}{*}{118.96} & 68.88  & 1.73$\times$                   \\
                          & Per-token dynamic       &                         & 88.55  & 1.34$\times$                   \\
                          & PTS (ours)              &                         & 69.30  & 1.72$\times$                   \\ \midrule
\multirow{3}{*}{VMamba-B} & Per-tensor static           & \multirow{3}{*}{165.18} & 89.93  & 1.84$\times$                   \\
                          & Per-token dynamic       &                         & 114.87 & 1.44$\times$                   \\
                          & PTS (ours)              &                         & 90.30  & 1.83$\times$                   \\ \midrule
\multirow{3}{*}{Vim-Ti}    & Per-tensor static          & \multirow{3}{*}{37.86}  & 32.58  & 1.16$\times$                   \\
                          & Per-token dynamic       &                         & 39.15  & 0.97$\times$                   \\
                          & PTS (ours)              &                         & 32.74  & 1.16$\times$                   \\ \midrule
\multirow{3}{*}{Vim-S}    & Per-tensor static          & \multirow{3}{*}{85.43}  & 67.75  & 1.26$\times$                   \\
                          & Per-token-dynamic       &                         & 75.67  & 1.13$\times$                   \\
                          & PTS (ours)              &                         & 67.86  &1.26$\times$                   \\ \midrule
\multirow{3}{*}{Vim-B}    & Per-tensor static          & \multirow{3}{*}{219.48} & 150.02 & 1.46$\times$                   \\
                          & Per-token dynamic       &                         & 169.4  & 1.30$\times$                   \\
                          & PTS (ours)              &                         & 157.03 & 1.40$\times$                  \\ \bottomrule
\end{tabular}
%}
\vspace{-2mm}
\caption{Latency comparison for Visual Mamba backbone. The batch size is 32.}
\vspace{-4mm}
\label{tab:Latency comparison}
\end{table}

\subsection{Computation Acceleration}
To demonstrate the efficiency of PTQ4VM from a hardware acceleration perspective, we implemented CUDA kernels and measured the execution latency. The kernels were built using CUTLASS \cite{kerr2017cutlass}, and all experiments were conducted on a single RTX 3090 with a batch size of 32.

\Cref{tab:Latency comparison} presents the end-to-end W4A4 latency when applying per-tensor static, per-token dynamic, and PTS quantization to the token dimension, with SmoothQuant applied by default. The results show that the proposed PTS achieves similar acceleration to per-tensor static across all models, while significantly outperforming per-tensor static in terms of accuracy, as seen in \Cref{tab:classification_results}. 
In particular, for VMamba-B, per-token dynamic quantization shows a 1.44$\times$ latency improvement compared to FP, while PTS achieves even greater acceleration at 1.83$\times$. 
These improvements are detailed in \cref{fig:profile-kernel-fp}, where PTS with W4A4 shows a significant improvement in the latency of the linear layer on VMamba-B, ultimately achieving a speedup of 1.83$\times$.
These results indicate that PTS can deliver optimal quantization quality with minimal overhead, comparable to the per-tensor static method. A comparison of accuracy between PTS and per-token dynamic is provided in \Cref{tab:dynamic_acc}.

\subsection{Ablation Study}
To evaluate the impact of each component in our proposed PTQ4VM, we conducted an ablation study (\Cref{tab:ablation_study}).

First, incorporating PTS into Vim-Ti allows us to account for token-wise variance, leading to notable accuracy improvements. Specifically, this method enables appropriate step size allocation to the crucial CLS token, proving particularly effective in models that use CLS tokens. As a result, we observe accuracy gains of 17.1\% at 6-bit and 1\% at 8-bit in Vim-Ti.

For VMamba, although there is no CLS token, accounting for token-wise variance still enhances accuracy at 6-bit and 4-bit levels. Additionally, by using a grid search to optimize the truncation level and minimize L2 loss at the layer level (labeled with +Truncation), both Vim and VMamba show significant accuracy improvements—approximately 40\% and 50\%, respectively, at 4-bit. This highlights the importance of proper truncation in static quantization. JLSS, designed to further optimize truncation, yields additional gains of 10.0\% and 7.3\% at Vim and VMamba, respectively.
% , while boosting accuracy by 10\% in Vim-T.
 
\begin{table}[t]
  \centering
  {\small{
  \begin{tabular}{ccc}
    \toprule
    \textbf{Granularity} & \textbf{Vim-S (\%)} & \textbf{VMamba-S (\%)} \\
    \midrule
    Per-tensor static & 0.2 & 4.5 \\
    Per-token dynamic & 72.6 & 82.4 \\
    PTS & 69.6 & 82.3 \\
    \bottomrule
  \end{tabular}
  }}
  \vspace{-2mm}
  \caption{Top-1 accuracy results of INT8 quantization across different activation granularity schemes.}
  \vspace{-2mm}
  \label{tab:dynamic_acc}
\end{table}

\begin{table}[t]
    \centering
    \small
    \setlength\tabcolsep{3pt}
    \begin{tabular}{cccccc}
    \toprule
    \multirow{2}{*}{\textbf{Model}} & \multirow{2}{*}{\textbf{Method}} & \multicolumn{4}{c}{\textbf{Top-1 Accuracy (\%)}} \\
                              &  & \textbf{FP16} & \textbf{W8A8} & \textbf{W6A6} & \textbf{W4A4} \\
    \midrule
    \multirow{4}{*}{Vim-Ti}
          & SmoothQuant& \multirow{4}{*}{76.1}& 74.7 & 54.5 & 0.1 \\
          & + PTS &  & 75.7 & 71.6 & 5.4 \\
          & + Truncation &  & 75.8 & 72.2 & 46.4 \\
          & + JLSS (ours) &  & \textbf{75.8} & \textbf{73.9} & \textbf{56.4} \\
    \midrule
    \multirow{4}{*}{VMamba-T}
          & SmoothQuant &   \multirow{4}{*}{82.6}& 82.6 & 81.8 & 1.7 \\
          & + PTS &  & 82.6 & 82.2 & 19.8 \\
          & + Truncation &  & 82.6 & 82.3 & 73.3 \\
          & + JLSS (ours) &  & \textbf{82.6} & \textbf{82.4} & \textbf{80.6} \\    \bottomrule
    \end{tabular}
    \vspace{-2mm}
    \caption{The effect of each component of PTQ4VM on the ImageNet top-1 validation accuracy.}
    \vspace{-5mm}
    \label{tab:ablation_study}
\end{table}

\section{Conclusion}
In this paper, we identified three key challenges that complicate quantization for Visual Mamba: (i) token-wise variance, (ii) channel-wise outliers, and (iii) the long tail of activation. To address these challenges, we propose PTQ4VM, a post-training quantization technique designed to tackle all three issues. Through Per-Token Static (PTS) Quantization, we effectively address token-wise variance, while the integration of SmoothQuant with PTS mitigates channel-wise outliers. Additionally, by utilizing JLSS, which jointly optimizes smoothing scale and step size, we achieve higher post-quantization quality. To the our best knowledge, this is the first quantization study on Visual Mamba. PTQ4VM can generates any quantized model within 15 minutes and is widely applicable to various Visual Mamba architectures across different downstream tasks, all while enabling faster inference. Our techniques offer up to a 1.83$\times$ speedup on real GPUs, broadening the practical application of Visual Mamba backbones.

\section*{Acknowledgement}
This work was supported by Institute of Information \& communications Technology Planning \& Evaluation (IITP) grant funded by the Korea government (MSIT) (RS-2019-II191906,  RS-2024-00396013, RS-2024-00457882).

%%%%%%%%% REFERENCES
{\small
\bibliographystyle{ieee_fullname}
% \bibliography{egbib}
\bibliography{arxiv}
}

\newpage
\appendix
\section{Implementation Details}
\subsection{Quantization Configurations}
In this section, we describe the detailed quantization configuration for reproducing PTQ4VM. The hyperparameters used in our image classification experiments are reported in \Cref{tab:quantization_hyperparams}. We denoted the learning rate for the smoothing scale $s$ as lr$_s$ and the learning rate for the quantization parameters (step size $\Delta_X$ and $\Delta_W$) as lr$_q$. Following SmoothQuant~\cite{xiao2023smoothquant}, we introduce the exponent value $\alpha$ used to initialize the smoothing scale as a hyperparameter. Here, $\alpha$ corresponds to the exponent term in \cref{eq:alpha}. In the case of \cref{eq:smooth_scale}, please note that $\alpha$ is set to 0.5.
\begin{equation}
\label{eq:alpha}
\mathbf{s} = \frac{\max(|\mathbf{X}|)^\alpha}{\max(|\mathbf{W}|)^{1 - \alpha}} \in \mathbb{R}^{D_{in}}.
\end{equation}
All models used in the experiments used pretrained checkpoints provided by the official repository of Vision Mamba~\cite{zhu2024vision}, VMamba~\cite{liu2024vMamba}, and LocalMamba~\cite{huang2024localMamba}.

\begin{table*}[h!]
    \centering
    \small  
    \resizebox{0.9\textwidth}{!}{%  
    \begin{tabular}{|c|cccc|cccc|cccc|cccc|cccc|}
    \toprule
    \multirow{2}{*}{Bits} & \multicolumn{4}{c|}{Vim-Ti} & \multicolumn{4}{c|}{Vim-S} & \multicolumn{4}{c|}{Vim-B} & \multicolumn{4}{c|}{VMamba-T} & \multicolumn{4}{c|}{VMamba-S} \\
    \cmidrule{2-21}
    & lr$_s$ & lr$_q$ & $\alpha$ & epoch & lr$_s$ & lr$_q$ & $\alpha$ & epoch & lr$_s$ & lr$_q$ & $\alpha$ & epoch & lr$_s$ & lr$_q$ & $\alpha$ & epoch & lr$_s$ & lr$_q$ & $\alpha$ & epoch \\
    \midrule
    8-bit & \multirow{3}{*}{1e-2} & \multirow{3}{*}{5e-4} & \multirow{3}{*}{0.5} & 10
    & \multirow{3}{*}{1e-3} & \multirow{3}{*}{5e-4} & \multirow{3}{*}{0.5} & 10
    & \multirow{3}{*}{1e-2} & \multirow{3}{*}{5e-4} & \multirow{3}{*}{0.5} & 10
    & 1e-4 & 1e-4 & 0.5 & 10 
    & 1e-4 & 1e-4 & 0.5 & 10 \\
    6-bit &  &  &  & 50
    &  &  & & 50
    &  &  & & 50
    & 1e-5 & 1e-5 & 0.5 & \multirow{2}{*}{50}
    & 1e-5 & 1e-5 & 0.5 & \multirow{2}{*}{50} \\
    4-bit &  &  &  & 100
    &  &  & & 100
    &  &  & & 100
    & 1e-4 & 1e-4 & 0.65 & 
    & 1e-4 & 1e-4 & 0.6 &  \\
    \bottomrule
    \end{tabular}%
    }
    
    \vspace{1mm}
    
    \small
    \resizebox{0.9\textwidth}{!}{%
    \begin{tabular}{|c|cccc|cccc|cccc|cccc|cccc|}
    \toprule
    \multirow{2}{*}{Bits} & \multicolumn{4}{c|}{LocalVim-T$^{\dagger}$} & \multicolumn{4}{c|}{LocalVim-T} & \multicolumn{4}{c|}{LocalVim-S}  & \multicolumn{4}{c|}{VMamba-B} & \multicolumn{4}{c|}{LocalVMamba-S} \\
    \cmidrule{2-21}
    & lr$_s$ & lr$_q$ & $\alpha$ & epoch & lr$_s$ & lr$_q$ & $\alpha$ & epoch & lr$_s$ & lr$_q$ & $\alpha$ & epoch & lr$_s$ & lr$_q$ & $\alpha$ & epoch & lr$_s$ & lr$_q$ & $\alpha$ & epoch \\
    \midrule
    8-bit & \multirow{3}{*}{1e-2} & \multirow{3}{*}{5e-4} & \multirow{3}{*}{0.5} & 10
    & \multirow{3}{*}{1e-2} & \multirow{3}{*}{5e-4} & \multirow{3}{*}{0.5} & 10
    & \multirow{3}{*}{1e-2} & \multirow{3}{*}{5e-4} & \multirow{3}{*}{0.5} & 10
    & \multirow{3}{*}{1e-4} & \multirow{3}{*}{1e-4} & 0.5 & 10
    & \multirow{3}{*}{1e-4} & \multirow{3}{*}{1e-4} & 0.5 & 10 \\
    6-bit &  &  & & 50
    &  &  & & 50
    &  &  & & 50
    &  &  & 0.5 & \multirow{2}{*}{50}
    &  &  & 0.5 & \multirow{2}{*}{50} \\
    4-bit &  &  & &100  
    & &  & & 100
    &  &  & & 100
    &  &  & 0.65 &  
    &  &  & 0.65 &  \\
    \bottomrule
    \end{tabular}%
    }
    \caption{Hyperparameters of different models for image classification across quantization settings.}
    \label{tab:quantization_hyperparams}
\end{table*}

For Object Detection and Instance Segmentation tasks, we used different learning rates and epochs depending on the quantization bit-width. For W8A8, we used a learning rate 1e-5 for both smoothing scale and step sizes, training for 10 epochs. For W6A6 and W4A4, we used a learning rate 1e-4 and trained for 50 epochs.
As mentioned in our main manuscript, we used the VMamba-T~\cite{liu2024vMamba} backbone trained with the MASK R-CNN 3X~\cite{he2017mask} MS setting, provided by the official VMamba repository. We applied quantization only to the backbone for our experiments. Additionally, we cropped the input images to 1280 $\times$ 800 to use PTS in our experiments.

\section{Acceleration Kernel}
In this section, we provide a more detailed explanation of the hardware implementation and experiment of PTQ4VM. 
\subsection{Implementation}
The hardware kernel of PTQ4VM is based on CUDA programming and CUTLASS 3.5~\cite{kerr2017cutlass}, and consists of four main parts: Activation smoothing, Quantization, INT GEMM, and Dequantization.

In the Activation smoothing part, we upload activation smoothing vector to the shared memory of each block in the Tensor core. We then apply element-wise multiplication to each row of the activation tensor to perform smoothing. We can utilize more efficient operation because we apply multiplication to shared memory uploaded data.

During the Quantization part, we quantize FP16 activations to INT4/INT8. Weights are not quantized during this phase because they are already saved and loaded as INT4/INT8 values. It is noticeable that the smallest data type of integer in PyTorch~\cite{NEURIPS2019_9015} is INT8. Therefore, when performing INT4 quantization, we concatenate two adjacent quantized values and pack them into INT8 data. The pseudo-code for this process is as follows: \resizebox{\linewidth}{!}{\texttt{packedData = (data[1] << 4) | (data[0] \& 15)}}\\

For the INT GEMM part, we utilize CUTLASS Tensor core INT4/INT8 GEMM, as CUTLASS is known to be the most efficient open-source linear algebra library currently available. In the Dequantize part, we generate the output step size tensor by conducting the inner product of the weight step-size vector and the activation step size vector. We employ CUBLAS GEMM for this operation because it is more efficient than CUTLASS for FP16 GEMM and is also used in PyTorch's default tensor multiplication, making implementation straightforward. The output step size tensor undergoes element-wise multiplication with the output tensor to produce the final result tensor. Similar to the activation smoothing process, this operation is performed with uploading to the shared memory of each tensor core block, thereby maximizing acceleration performance.

\begin{figure}
\centering
\includegraphics[width=0.9\linewidth]{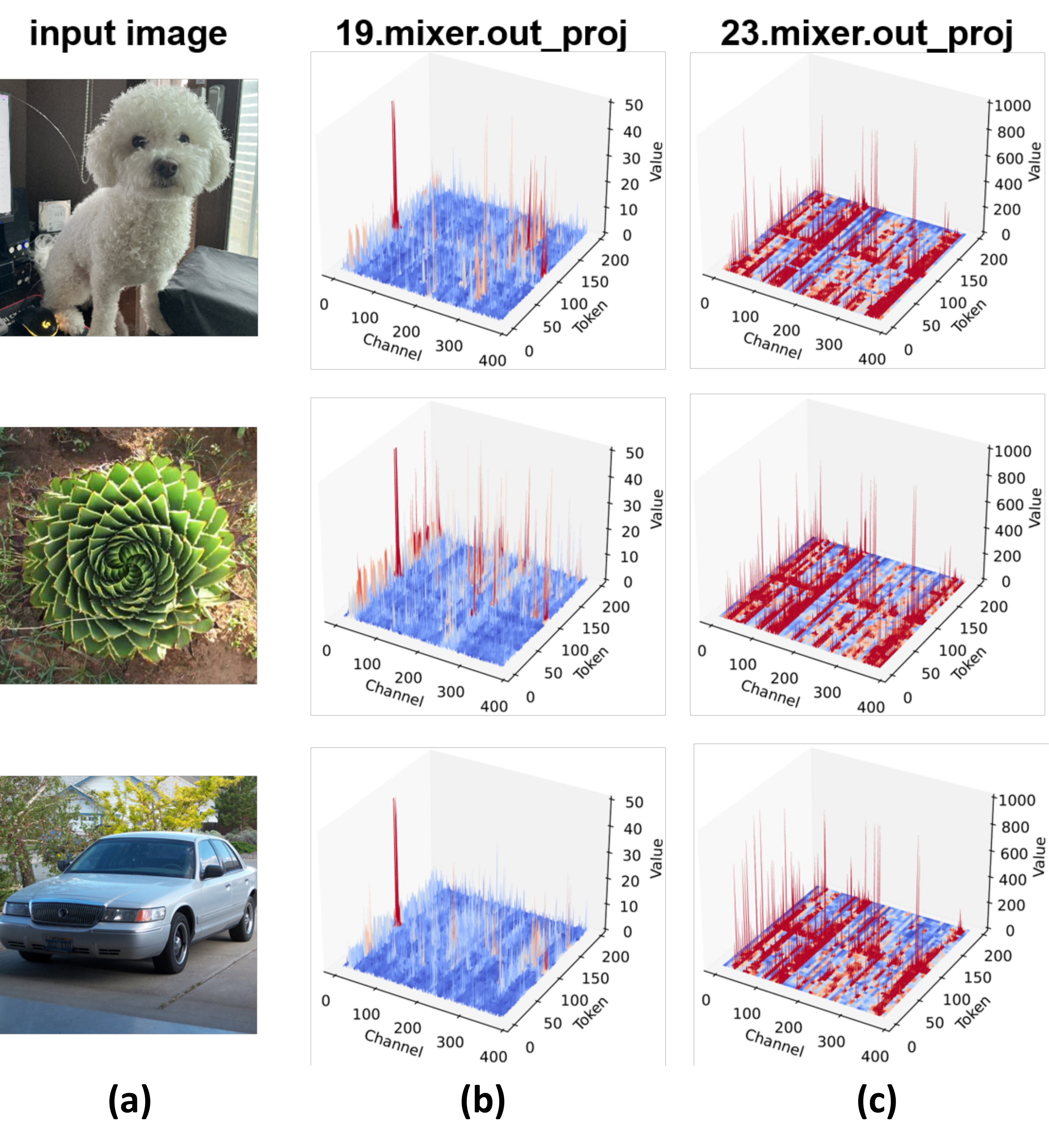}
\caption{The activation distribution of the out\_proj layer in Vim-Ti. (a) the input image, (b) the 19th out\_proj, and (c) the 23rd out\_proj. Here, the layer index uses 0-based numbering.}
\label{fig:vim_t_other_layer}
\end{figure}

\subsection{Experimental Settings}
The latency measurement for the kernel-implemented model were conducted on a single RTX 3090 GPU with a batch size of 32. To ensure accurate measurements, we employed a warming-up phase consisting of 100 iterations prior to the actual timing. Subsequently, we measured the 100 times of inference latency and reported the median value. By this methodology, we can conduct more reliable assessment of the performance by mitigating the effects of initial overhead and potential outliers. The use of the median value provides a robust tendency that is less sensitive to extreme values compared to the arithmetic mean.

\section{Additional Visualization}

\begin{figure*}
\centering
\includegraphics[width=0.95\linewidth]{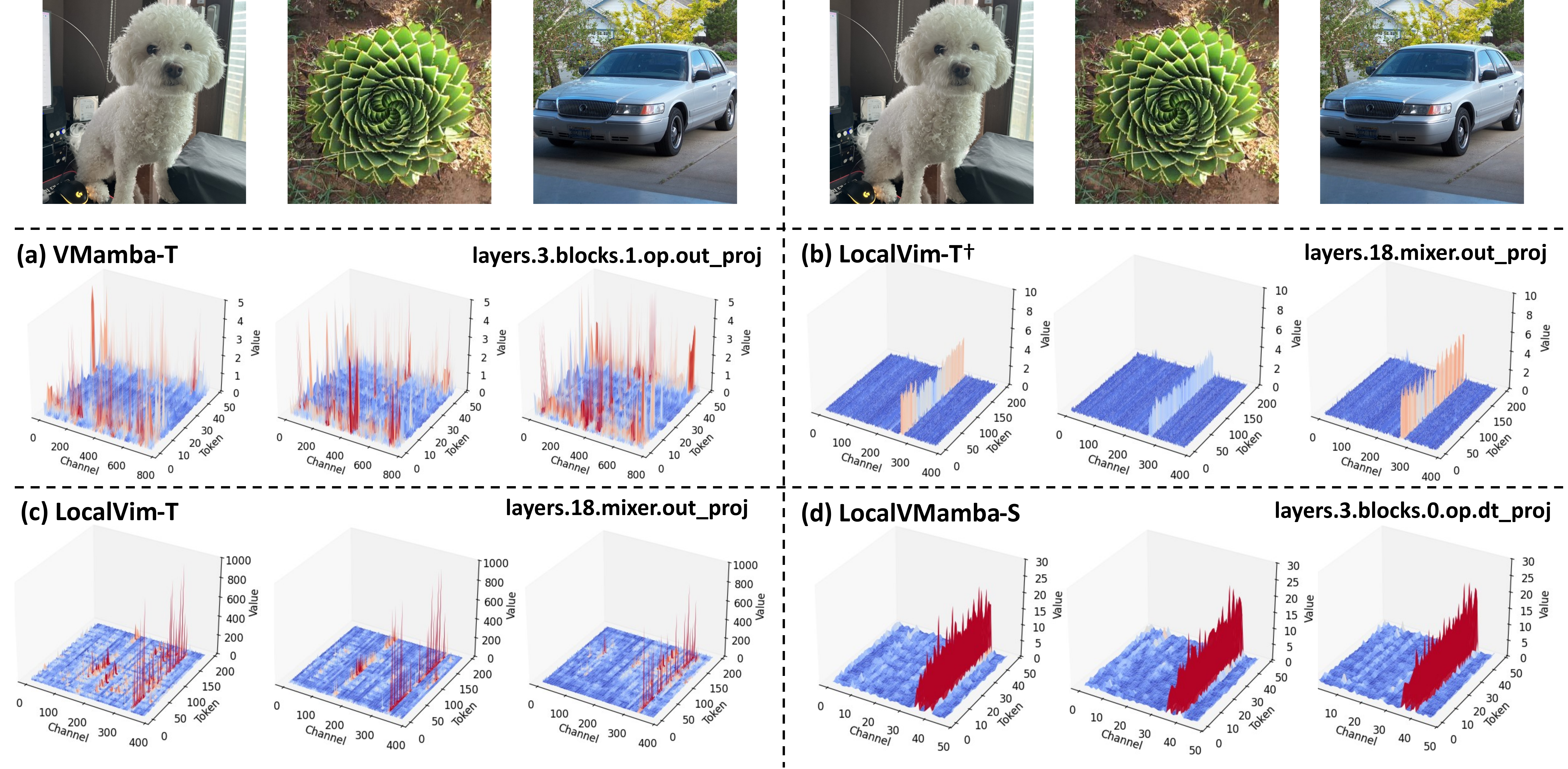}
\caption{Activation distributions for different input images. The x-axis is the channel dimension, the y-axis is the token dimension, and the z-axis represents the absolute value. (a) VMamba-T, (b) LocalVim-T$^\dagger$, (c) LocalVim-T, and (d) LocalVMamba-S.}
\label{fig:observation1_2}
\end{figure*}

\begin{figure*}
\centering
\includegraphics[width=0.95\linewidth]{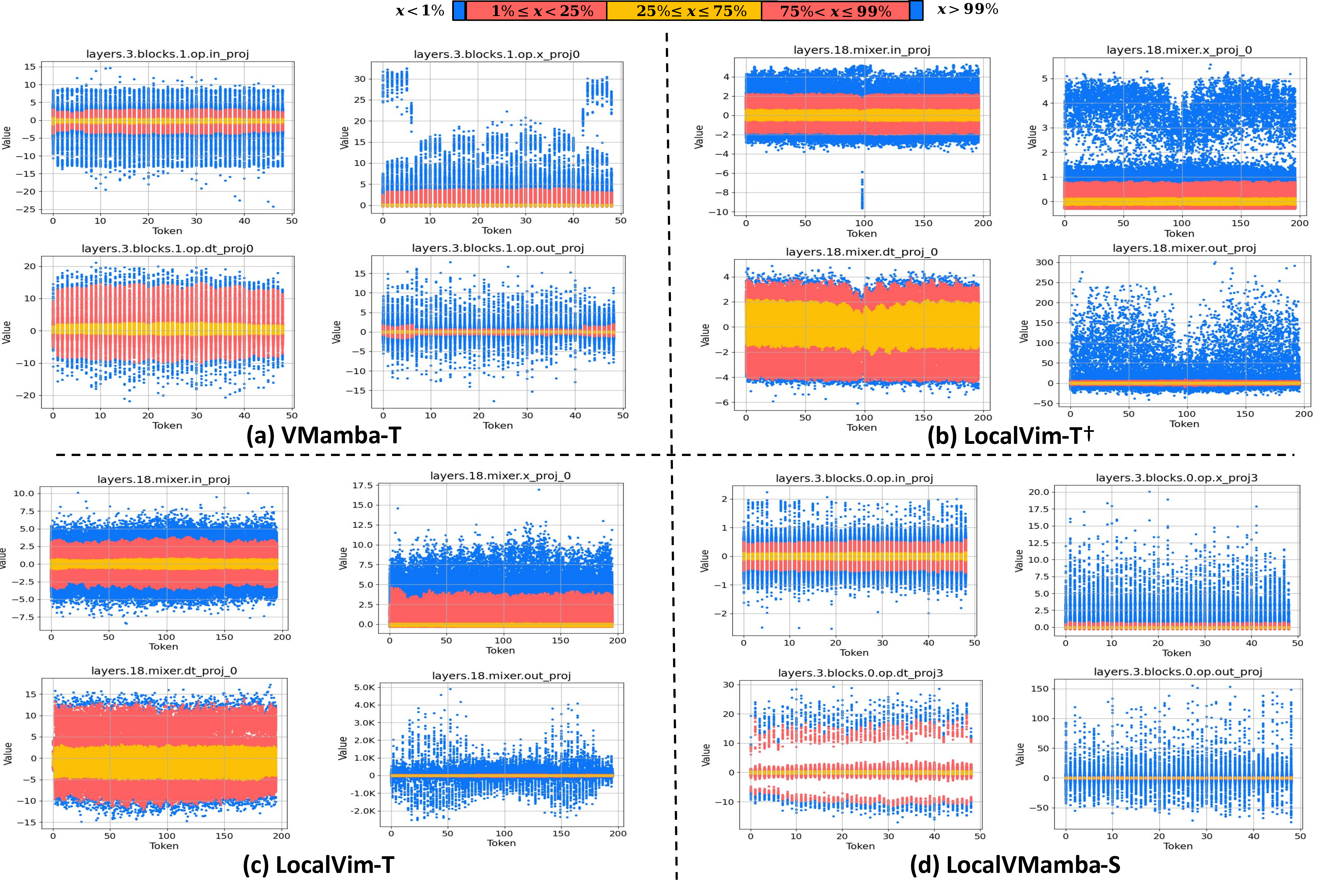}
\caption{Activation distributions across token direction. We visualized the above distribution using 32 randomly sampled images. (a) VMamba-T, (b) LocalVim-T$^\dagger$, (c) LocalVim-T and (d) LocalVMamba-S.}
\label{fig:obasevation3}
\end{figure*}

In the main manuscript, we only reported the distribution of the 22nd out\_proj layer of Vim-Ti. These observations are also present at other layer indices, and additional visualizations can be found in \cref{fig:vim_t_other_layer}. Furthermore, we have reported Observations 1 and 2 for VMamba, LocalVim$^\dagger$, LocalVim, and LocalVMamba in \cref{fig:observation1_2}, and Observation 3 in \cref{fig:obasevation3}. We visualized the input activation of the out\_proj layer of backbones in \cref{fig:observation1_2}, except for LocalVMamba-S case, which is visualization of dt\_proj layer.

%-------------------------------------------------------------------------
\end{document}